# The Role, Trends, and Applications of Machine Learning in Undersea Communication: A Bangladesh Perspective

Yousuf Islam(✉)[1], Sumon Chandra Das(✉)[1], Md. Jalal Uddin Chowdhury(✉)[1,2]

1 Department of Computer Science and Engineering, Leading University, Sylhet 3112, Bangladesh
2 DeepNet Research and Development Lab, Sylhet 3100, Bangladesh

**Abstract** The rapid evolution of machine learning (ML) has brought about groundbreaking developments in numerous industries, not the least of which is in the area of undersea communication. This domain is critical for applications like ocean exploration, environmental monitoring, resource management, and national security. Bangladesh, a maritime nation with abundant resources in the Bay of Bengal, can harness the immense potential of ML to tackle the unprecedented challenges associated with underwater communication. Beyond that, environmental conditions are unique to the region: in addition to signal attenuation, multipath propagation, noise interference, and limited bandwidth. In this study, we address the necessity to bring ML into communication via undersea; it investigates the latest technologies under the domain of ML in that respect, such as deep learning and reinforcement learning, especially concentrating on Bangladesh scenarios in the sense of implementation. This paper offers a contextualized regional perspective by incorporating region-specific needs, case studies, and recent research to propose a roadmap for deploying ML-driven solutions to improve safety at sea, promote sustainable resource use, and enhance disaster response systems. This research ultimately highlights the promise of ML-powered solutions for transforming undersea communication, leading to more efficient and cost-effective technologies that subsequently contribute to both economic growth and environmental sustainability.

## 1 Introduction

Modern marine operations such as oceanographic research, defense as well as industrial exploration, and disaster response all depend on underwater communication. While electromagnetic (EM) waves dominate terrestrial communication systems, acoustic signals have become a widely used means of underwater communication because of the severe attenuation of EM waves in water. Acoustic signaling is the better option in underwater scenarios; nevertheless, it still has drawbacks such as high signal loss, multi-path channels, noise, and limited bandwidth [1, 2]. These challenges are further compounded by the dynamic nature of the marine environment including salinity gradients, temperature variations, and sedimentation [3]. One area in which this complexity is evident is the Bay of Bengal, whose waters are vital to Bangladesh. A vast ecological feature and resource, its broad Exclusive Economic Zone (EEZ) contains valuable fisheries, hydrocarbons, and important diversity. The strategic development of a robust undersea communication network for this region is essential for sustainable resource use, maritime security, and disaster readiness, among other national interests [4]. Solutions to overcome such problems would need to combine state-of-the-art techniques in signal processing and regional issues so that they are appropriately relevant for cooperation in the respective regions under the relevant environmental conditions.

And yet undersea communication has fundamental performance and reliability questions that need to be addressed. The traditional acoustic communication system suffers from low data rates, and long transmission dis-tances, and is easily affected by environmental noise [5]. These challenges are even greater for areas like the Bay of Bengal in which high sedimentation, salinities, and ocean currents affect signal degradation [6]. Moreover, conventional communication technologies are often inadequate to meet the growing demands of modern maritime activities that require high-speed data exchanges, including underwater robotics, remote monitoring of marine life and environmental conditions, and disaster response [7]. ML has become one of the most impressive solutions to solve many complex engineering problems, including underwater communication. However, its use in the context of Bangladesh's maritime domain is limited. This gap limits the country's capacity to capitalize on its marine resources and respond to various environmental and disaster challenges [8]. To overcome this gap, it is essential to develop optimized ML-based solutions, which will not only improve the resilience, flexibility, and efficacy of underwater communication networks but also address Bangladesh's special environmental and socio-economic situation.

Over the years, underwater communication technology has been developed with the use of frequency shift keying (FSK) and phase shift keying (PSK) techniques, and more advanced technologies such as orthogonal frequency division multiplexing (OFDM) [9, 10]. Nonetheless, these approaches often fall short when dealing with the complexities of noisy and dynamic underwater environments, like those found in the Bay of Bengal. For example, high ambient noise and multipath propagation can introduce serious signal degradation, causing a degradation in the reliability of traditional communication methods [11]. These challenges necessitate the use of ML-based techniques, which have shown plenty of promise in various fields including signal processing, denoising, and predictive modeling. By improving the reliability and efficient usage of underwater communication, ML algorithms provide new tools to optimize communication protocols, channel estimation, and error correction techniques [12]. While improvements in these areas continue, most studies address generalized applications and do not consider regional factors, such as the conditions in Bangladesh. Additionally, the combination of environmental and ecological factors in ML approaches for underwater communication is still a relatively underinvestigated field. Addressing this gap underscores the need for more targeted research that connects global innovations with local applications to high-growth potential regions such as the Bay of Bengal. In this paper, we aim to fill the gaps by covering the trends and applications of machine learning in undersea communication and identifying the gaps with respect to the unique challenges of the Bay of Bengal. The study aims specifically to:

- Summarize the state-of-the-art ML techniques applied in underwater communication, outlining notable trends, challenges, and research opportunities in this domain.
- Create a tailored framework that harmonizes ML to Bangladesh's maritime infrastructure, accounting for technical and ecological implications.
- Emphasize the importance of economic development and simultaneously achieving environmental sustainability so that ML-based solutions can fit in with Bangladesh's strategic interests in the Bay of Bengal.

This research contributes to the establishment of sustainable and efficient undersea communication systems that address technological and ecological priorities by applying state-of-the-art ML techniques to the specific challenges of the maritime sector in Bangladesh.

## 2 Background Study

Maritime operations heavily rely on undersea communication, but it is often challenged by factors such as signal fading, noise, and multipath. Reconfigurable networks, on the other hand, face challenges—such as changing dynamics—hence impeding reliable communication. ML enables innovative aspects in underwater networks such as effective signal processing, optimizing transmission efficiency, and advanced real-time decision-making.

### 2.1 Overview of Undersea Communication

Undersea communication is a specialized field of telecommunication, allowing information transfer in water environments. Acoustic signals are used almost exclusively for these underwater communications because electromagnetic waves do not propagate well through water media since they decay and scatter exponentially. Acoustic communication is extensively used in numerous applications in underwater robotics, maritime navigation, environmental monitoring, and offshore exploration [13]. Although the use of acoustic waves is considered suitable for long-range communication, there are several limitations associated with it, such as bandwidth limitation, signal loss, multipath effect, noise interference, etc. Here these constraints affect the efficacy and reliability of underwater communication systems [14]. Signal attenuation, for example, reduces the strength of acoustic signals over larger distances, while multipath propagation—based on reflections from the seabed, water surface, and other objects—adds to signal distortion. Additionally, ambient noise caused by marine life, ship traffic, and the environment, all represent significant challenges for signal clarity [15].

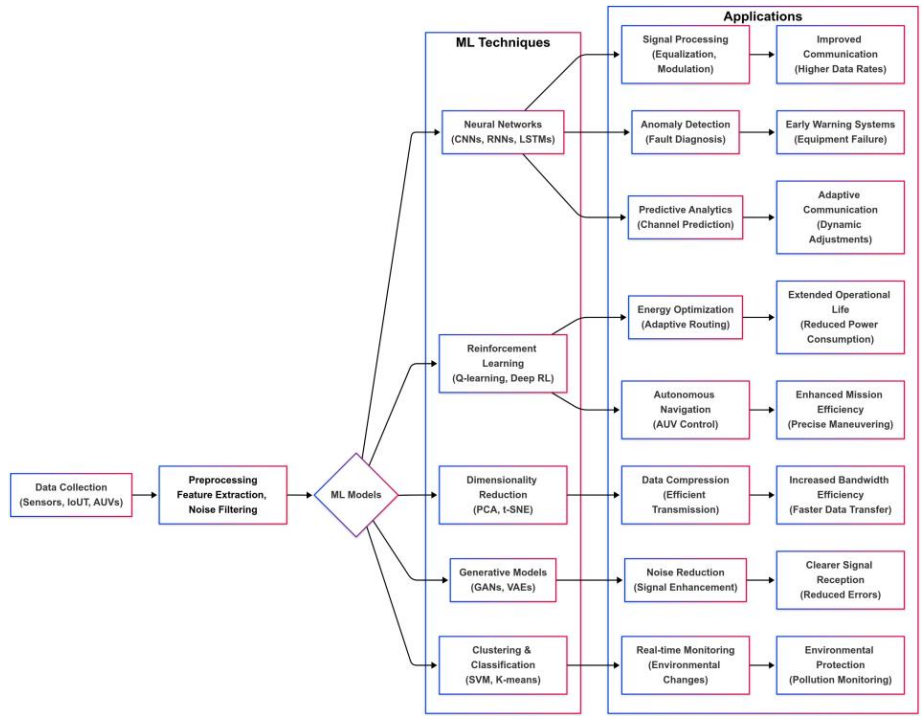

**Fig. 1** ML in Undersea Communication – Conceptual Diagram

The challenges are compounded in the Bay of Bengal where unique environmental conditions like salinity gradients, temperature differentials as well as seasonal oceanographic processes must also be taken into consideration. As a strategically important region, the need for effective undersea communications in the area is essential for ensuring maritime safety, conservation response systems, and sustained development. Acoustic methods technologies provide solutions to these problems, however, they are only half of the equation.

### 2.2 Machine Learning in Communication

ML has been adopted as a game-changing technology by communication systems, and solutions for dynamics adaptation to environmental variations and optimization of system performance have been proposed. Manual configurations or rule-based approaches are used to address the issues of noise interference, resource allocation, and signal distortion in traditional communication [16]. Conversely, ML offers data-driven approaches that can learn from the patterns of the data and improve the performance of the system in real-time. Fig. 1 presents a conceptual framework for ML applications for undersea communication, illustrating the general flow from data generation, through processing and eventual deployment at real-world applications. ML methods are transforming multiple elements of underwater communication: Signal processing and noise reduction: Signal processing refers to the analysis and manipulation of signals, whereas ML techniques like RNNs and Autoencoders help in denoising and retrieving lost signals [17]. Models like these are most useful in areas accompanied by high background noise, such as the Bay of Bengal.

Channel Modeling and Prediction: ML models analyze environmental data like salinity and temperature to predict channel behavior and also assist in optimizing the transmission parameters [18]. Adaptive modulation and coding are further aided by reinforcement learning. Data Error Detection and Correction: By identifying error patterns and making suggestions for corrective actions, neural networks improve the reliability of communication systems. These techniques allow accurate data transmission in the presence of challenging conditions [19]. Energy Optimization: ML-based resource allocation algorithms are also designed to minimize energy [20] usage which is a key consideration for autonomous underwater devices powered by limited supply.

Internationally, countries such as Japan and the United States have already incorporated ML into their underwater communication networks with considerable improvements in efficiency and reliability. In Bangladesh, however, analogous applications are at a nascent stage, creating a tremendous potential for expansion and creativity.

### 2.3 Bangladesh's Maritime Context

Bangladesh is on the territory of a 1,83,613 square kilometers Exclusive Economic Zone (EEZ) which has rich marine resources, and the Bay of Bengal is an area of great economic and strategic importance for Bangladesh [21]. In particular, fisheries are a national economic powerhouse, and offshore hydrocarbon reserves remain a newly accessible resource with unlimited potential for energy independence. In addition to resource exploitation, the susceptibility of the region to cyclones and other natural disasters underscores the importance of strong communication infrastructure to aid in disaster preparedness and response [22].

But its intricate underwater terrain — with sedimentation, turbulent water flows, and changing salinity rates — represents major hurdles for traditional communications systems. Even though it is not a permanent solution, implementing ML technologies in underwater communication networks could help in improving these limitations. ML demonstrates its transformative potential in enhancing maritime operations in Bangladesh through applications including real-time monitoring of fish stocks, pollution detection, and natural disaster early warning systems that benefit not only organizations and SDG progress but also other sectors across the country.

## 3 Role of Machine Learning in Undersea Communication

Applying ML to enable undersea communication helps to redefine how data are communicated, transmitted, computed, and interpreted in difficult aquatic realms. Models powered by ML improve the efficiency and reliability of communication via noise reduction, channel prediction, and adaptive modulation. In this section, we will focus on how the use of ML techniques can overcome the primary limitations of traditional underwater networks thus paving the way for smarter and robust systems. Also, table 1 provides a comparative evaluation of different ML methods employed in undersea communication, emphasizing their benefits and limitations.

### 3.1 Signal Processing and Noise Reduction

Communication in the underwater environment is highly disturbing due to various ambient noises such as from marine life, human activities, and environmental conditions [17]. However, these problems can be alleviated by using more advanced techniques provided by the ML community, ensuring more reliable and high-performance signal processing. ML models like Recurrent Neural Networks (RNNs) and Denoising Autoencoders are trained to recognize patterns from a noisy dataset, thereby separating the meaningful signal from the interference. In the Bay of Bengal, where ambient noise levels are highly variable, the application of ML-driven noise reduction techniques shows promise in improving the clarity of acoustic communication for applications, such as environmental monitoring and resource management [23].

In addition to filtering noise, ML algorithms provide real-time signal enhancement through the reconstruction of mutilated or incomplete data. Using technology such as Generative Adversarial Networks (GANs), our signals can be restored to their original quality, allowing for reliable communication over difficult channels at the expense of minimal bandwidth [24]. These developments have particular significance both for disaster response and maritime safety in Bangladesh, where uninterrupted and clear communication is essential for success.

### 3.2 Channel Modeling and Prediction

Therefore, the modeling of underwater channels is critical to the optimization of data transmission. ML models provide a big data-based alternative to traditional approaches, which often struggle to react to constantly changing aquatic environments. These models utilize the relationships between environmental variables (salinity, temperature, and pressure) to predict the future behavior of the channel and prescribe the best possible transmission parameters based on the prediction. By predicting the channel state based on previous experience, reinforcement learning enhances the aforementioned process even further by adaptively altering modulation schemes and transmission power according to real-time feedback [18].

ML-driven channel modeling in Bangladesh where the Bay of Bengal faces regular environmental changes has crucial potential to enhance communication reliability. For example, ML guided adaptive strategies can also reduce the diversity of the effects of seasonal salinity variations or sedimentation on the propagation of signals, thereby providing robust operation in a variety of applica-tions from the sea [25].

### 3.3 Data Compression and Transmission

Undersea communication systems face practical challenges in terms of limited bandwidth and high latency, making data compression and transmission an important aspect of efficient execution. Autoencoders and Principal Component Analysis (PCA) gain insights from ML methods compressing data by removing non-informative redundancy and obtaining necessary features. These methods enable critical information to be passed through without overloading the network [26].

In underwater acoustic networks, dynamic adjustments of routing strategies and prioritization of essential data with respect to delaying and losing the data packets enable accident avoidance in data transmission making the reinforcement learning algorithms more applicable to improving the reliability of the transmission. Utilizing channel conditions in a real-time manner, predictive ML models could choose the best transmission parameters, thereby reducing errors and retransmissions [27]. Through the Bay of Bengal, these approaches guarantee stable communications for monitoring disasters as well as marine study, regardless of the regularly shifting ecological conditions.

### 3.4 Anomaly Detection and Environmental Adaptation

Robust underwater communication systems require effective environmental change detection and adaptation. Examples of ML models, like Long short-term memory (LSTM) networks and clustering algorithms, detect anomalies in data and output how to take action to ensure system reliability. An ML model can e.g. identify sudden fluctuations in salinity or temperature possibly caused by underwater disturbances or the presence of signal interference [28]. Through the Bay of Bengal and environmental inter-condition from seasonal to tidal, anomaly detection plays a vital role in the effective operation of communication systems. Reinforcement learning models dynamically adapt parameters like communication power levels and modulation schemes according to environmental feedback in real-time conditions [29]. This adaptability allows for consistent performance, supporting applications including disaster response, fisheries management, and environmental monitoring.

## 4 Emerging Trends in Undersea Communication

ML advances are generating innovations for undersea communication by using autonomous decision-making to improve data processing. The use of ML with different new technologies such as the Internet of Underwater Things (IoUT), edge computing, and hybrid communication systems are making the systems more efficient and adaptive. This segment analyzes the emerging trends influencing the future of underwater networks and the implications they hold for maritime operations. Figure 2 depicts the structure of an IoUT network that incorporates ML-based decision-making, enhancing the efficiency of data collection and transmission in underwater settings.

### 4.1 Integration with the Internet of Underwater Things (IoUT)

The IoUT represents a new wave of undersea communication. IoUT allows real-time data transmission between hundreds of connected underwater devices, for example, sensors, AUVs, and buoys [30]. This is where ML comes as a big factor to improve the effectiveness of IoUT systems in the field of intelligent decision-making, data processing, and resource optimization [31]. One of the most potential applications of IoUT in Bangladesh can be Environmental monitoring in IoUT with ingredients of ML. For instance, IoUT-enabled sensors can monitor water quality, salinity, and pollution in the Bay of Bengal to generate actionable insights for sustainable fisheries and marine biodiversity conservation [32]. Likewise, such ML-powered IoUT systems can forecast and minimize the impacts of natural disasters by predicting seismic and oceanographic data, which ultimately leads to timely efficient responses.

### 4.2 Autonomous Underwater Vehicles (AUVs)

However, Artificial Intelligence (AI) decision-making capabilities allow Autonomous Underwater Vehicles (AUVs) to revolutionize undersea operations and exploration. These autonomous systems are able to navigate complex underwater settings, capture high-resolution information, and complete complex tasks with limited human supervision. AUVs play a significant role in hydrocarbon exploration, marine research, surveillance, and so forth, in Bangladesh [33]. By providing them with advanced learning techniques, it is possible to enhance their functionality in adaptive path planning, obstacle avoidance, and energy-efficient operations. For example, Reinforcement learning can be used by AUVs deployed in the Bay of Bengal to plan their routes in a way that ensures coverage of large maritime areas while minimizing energy consumption. Moreover, AUVs with ML-enabled anomaly detection systems are capable of detecting underwater obstacles, structural damages, or illegal activities [34], benefiting maritime security and operational efficiency runtime.

**Table 1** Machine Learning Techniques and Their Applications in Undersea Communication

| ML Technique | Application | Key Advantage | Limitation |
|---|---|---|---|
| CNNs & RNNs | Noise Reduction & Signal Processing | Improves visibility in noisy environments. | High computational cost |
| Reinforcement Learning | Channel Optimization | Adapts dynamically to changes in the environment. | Requires extensive training data |
| Autoencoders | Data Compression & Transmission | Decreases data traffic in limited networks. | Loss of minor details in compression |
| GANs | Signal Restoration | Recover degraded acoustic signals effectively | Demands high computing power |

### 4.3 Hybrid Communication Systems

Acoustic, optical, and radio frequency (RF) technologies are capable of being adopted as standalone methodologies, with each having merit, however, there are potential limitations to use as described below, leading to the proposed concept of hybrid communication systems in underwater environments. The combination exploits the advantages of every modality: acoustic signals can provide long-range communication, optical signals deliver high-speed data rates, and RF signals are more suitable for surface communication, complementing one another in this regard [35]. ML helps in improving the performance of hybrid systems by determining which is the most suitable approach to communicate according to environmental factors.

In cases, such as in the Bay of Bengal, where salinity and turbidity levels differ, ML techniques can study real-time water body properties to choose an optimal method of transmission for uninterrupted operation. They facilitate various applications including environmental monitoring, disaster response, and resource exploration by enabling a combination of different communication technologies. ML-powered hybrid systems can effectively address the majority of hurdles related to underwater communication in Bangladesh, thus permitting sustainable maritime operations in the region for enhanced decision-making [36].

### 4.4 Advancements in Edge Computing

Edge Computing is still a relatively nascent yet transformative technology that brings the processing of data closer to the source (undersea data) as opposed to relying on centralized servers [37]. This paradigm effectively diminishes latency by conserving bandwidth and improving the real-time processing efficiency of underwater data. The combination of ML algorithms with edge computing devices allows underwater systems to make decisions autonomously and process large amounts of data locally, even in resource-constrained settings [38].

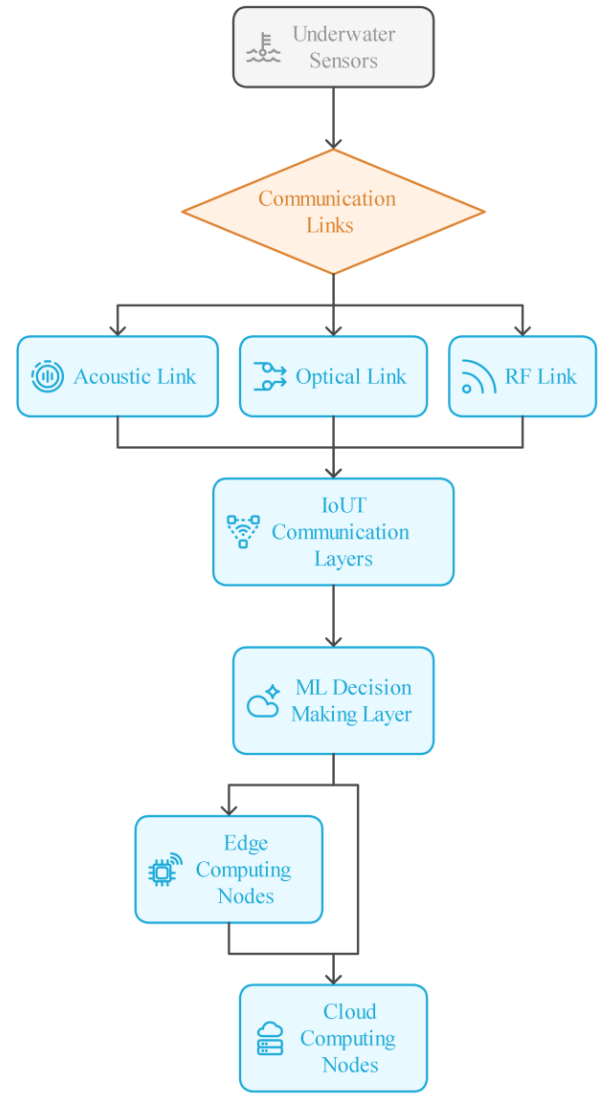

**Fig. 2** IoUT Architecture with Machine Learning Integration

Edge computing has potential applications for environmental monitoring, disaster management, and maritime security, specifically in the context of the Bay of Bengal. As an example, sensors deployed underwater with edge computing capabilities can process the salinity of the water, the temperature, the levels of pollution, etc., by analyzing the data in the local area themselves and thus only reporting advanced knowledge to the core system [39]. This approach results in lower communication overhead, timely analysis of data, and longer operational lifespan of devices deployed underwater as it saves energy.

ML can be used to augment edge computing to provide smart decision-making capabilities that can help you to analyze data and learn from it. An example is AUVs with edge processors that can facilitate navigation within complex marine environments, processing environmental data on-site at the edge, and avoiding hazards, while optimizing their routes [40]. This prerequisite becomes highly useful in cases such as disaster response where immediate action is essential.

Potential applications for edge computing in underwater environments range from ocean mapping to environmental monitoring to autonomous underwater vehicles (AUVs), though its adoption is hindered by challenges such as energy constraints and the demands for robust miniaturized hardware to survive extreme underwater environments. However, overcoming these challenges with advances in low-power computing and resilient device design will be crucial to unlock the full promise of edge computing for undersea communication.

## 5 Recommendations for Implementation

A systematic method is crucial for effectively adopting ML into undersea communication systems and environments as challenging as the Bay of Bengal. The experience also suggests several strategic steps to achieve such integration:

### 5.1 Invest in Data Collection and Preprocessing Infrastructure

This is mostly because well-performing ML models depend on having large, top-tier datasets for training and validation. Data collection infrastructure such as underwater sensors, IoUT nodes, and AUVs must be prioritized for deployment all over the coast of Bangladesh in order to gather environmental and operational data [41]. Needless to say, investing in the necessary preprocessing tools to clean, annotate, and standardize data is just as critical for improving model accuracy and reliability.

### 5.2 Promote Multidisciplinary Research and Collaboration

With ML in undersea communication positive applications, its inherently multidisciplinary nature requires individuals from marine biology, environment science, and computer engineering. Such collaborations will expose Bangladesh to global best practices, innovative technologies, and funding, in partnership with global research institutions and undersea systems experts [42].

### 5.3 Develop Context-Specific ML Models

Generic ML models are indiscriminate thus failing to consider the region-specific environmental and operational conditions such as the salinity gradients prevalent in Bay of Bengal and sedimentation patterns. Such ML model tuning will be useful to improve their performance and address the specific needs in the region. Supervised learning algorithms developed for predicting seasonal variations in water conditions would be helpful for ensuring communicating reliability for example [43].

### 5.4 Adopt Modular and Scalable System Designs

When integrating ML technologies, it should be with focus on a modular and scalable approach. Using modular systems also means that devices such as data collection devices or processing units can be upgraded independently without having to re-engineer new technologies into the entire network. Scalability enables infrastructure to be adapted for future use cases, where ML technologies become more advanced and operational needs increase [44].

### 5.5 Focus on Energy Efficiency

A major challenge in undersea operations is energy limitation. Our task is to create power-efficient ML algorithms and hardware solutions for underwater devices. During this period, the development of edge computing and energy harvesting methods (e.g., harvesting energy from underwater currents or thermal gradients) would enable the more prolonged use of underwater systems [45].

## 5.6 Establish Regulatory and Ethical Guidelines

For the deployment of ML-enabled solutions to be sustainable and ethical in Bangladesh, specific regulatory frameworks should be enacted addressing data privacy, environmental sustainability, and operational safety. These guidelines should be in accordance with global standards while taking into account region-specific challenges and priorities [46].

## 5.7 Provide Capacity Building and Skill Development

An ML-integrated system must be deployed and maintained as successfully as possible, which requires a skilled workforce. These include the development of educational programs and specialized training courses on marine technology, artificial intelligence, and underwater communication [47]. Moreover, public-private partnerships can amplify such capacity-building efforts by offering states valuable access to advanced systems and tools.

# 6 Comparative Case Studies and Global Insights

ML is being successfully used in undersea communication in many parts of the world, and lessons can be learned from these applications for Banglad- esh to consider while tackling the challenges in the Bay of Bengal. By studying effective implementations of ML in other countries like Japan, the US, and Norway, this part shows how advanced ML methods can enhance the efficiency of underwater communication, environmental monitoring, and resource management [48]. Table 2 outlines the significant machine learning advancements in undersea communication based on global case studies, offering perspectives for Bangladesh's implementation of comparable technologies.

## 6.1 Japan's Advanced AUV Deployment for Marine Research

Japan was the first to explore and manage its vast marine resources using Autonomous Underwater Vehicles powered by ML. Reinforcement learning algorithms enable these AUVs to effectively navigate through complex underwater environments. They gather high-resolution data for use in disaster mitigation, ocean mapping, and resource exploration. An important example of such applications is in post-tsunami assessments where AUVs offer valuable information about underwater land shifts and structural damages [49].

Now, these AUVs also employ adaptive path planning and anomaly detection systems to enhance their performance. Using ML algorithms that dynamically adjust their processing of environmental data, the AUVs can avoid hazards such as submerged debris or damaged underwater infrastructure. By employing energy-efficient edge computing hardware, Japan has been able to extend the operational time of these AUVs, further minimizing maintenance operations or retrieval [50]. Adapting similar AUV technologies has a potential to change the landscape of marine research and hydrocarbon exploration in the Bay of Bengal for Bangladesh [51]. ML-driven navigation systems would enable these AUVs to efficiently navigate through complex sedimentation areas while minimizing energy use, a possibility within the large and dynamic underwater habitats of the region.

## 6.2 The United States' Sensor Networks for Real-Time Monitoring

The United States has successfully realized its ubiquitous underwater sensing in the Pacific Ocean, which is tailored for disaster response, environmental monitoring, and maritime security [52]. ML algorithms are leveraged in these networks for real-time signal processing, predictive analytics, and for anomaly detection. For example, ML-empowered sensors have the facilities to filter out the background noise generated by marine creatures and anthropogenic activities so that the integrity of transmitted data can be preserved.

A major use case is integrating predictive analytics to process seismic and oceanographic data. For example, these models offer tsunami early warning, allowing for evacuations and the preparation of disaster response plans [53]. Doctor-authored content: The integration of edge computing within sensor networks enables processing data at nearby locations, drastically reducing latency and ensuring timely responses in the event of an emergency.

Bangladesh, for example, might deploy similar sensor networks to measure salinity, water pollution, and dynamics of fish stocks in the Bay of Bengal [54]. The region's susceptibility to cyclones and tidal surges also emphasizes the need for advance warning systems. Integrating the ML-driven sensors with the existing disaster management framework of the region, Bangladesh could considerably mitigate the repercussions of the calamities and thereby protect her coastal communities.

### 6.3 Norway's IoUT-Powered Smart Fisheries Management

By applying the Internet of Underwater Things (IoUT) and ML technologies in fisheries management, Norway has evolved into a global leader in sustainable fisheries management. Norway tracks the movement of fish, water quality, and temperature changes in real-time by using a network of underwater devices [55]. Unsupervised learning algorithms are utilized by these systems to study and gain valuable information about the habitats and population dynamics of marine species through their behavioral patterns.

The IoUT devices driven by ML, predict the dynamics of fish stock that helps fishermen avoid over-harvesting while preserving marine biodiversity by scheduling the harvesting accordingly. This strategy has improved the sustainability of Norway's fisheries over the long term while also preserving the environment. In Bangladesh, with fisheries being a key economic sector, IoUT systems would redefine resource management. For example, ML-based predictions about the patterns of migration and breeding of fish could be used by fishing facilities to increase their yield without compromising sustainability [56]. This, in turn, would mean real-time monitoring of the quality of water in the Bay of Bengal and pollution levels to preserve marine diversity.

### 6.4 Key Lessons for Bangladesh

Such applications showcase the transformative power of ML in undersea communication and maritime operations. Japan's AUV systems demonstrate the necessity of autonomous navigation and energy efficiency for long-term monitoring; the United States' sensor networks show the potential of real-time data processing and early warning for disaster prevention. ML can even increase sustainability and resource management as demonstrated by Norway's IoUT applications. With an adaptation of these strategies to its specific environmental and socio-economic context, Bangladesh will be able to realize the full potential of ML technologies in marine research, disaster response, and sustainable development [57]. However adopting these systems involves tackling technical, financial, and regulatory hurdles, highlighting the need for tailored solutions and collective action.

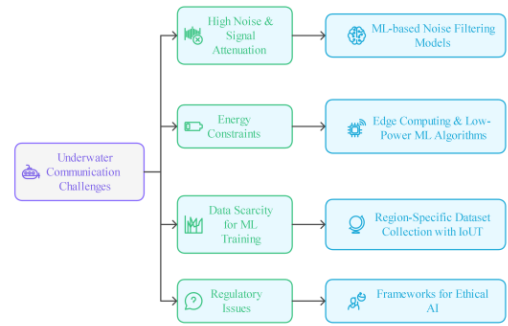

**Fig. 3** Challenges and Solutions in ML-Based Undersea Communication

## 7 Discussion and Analysis of Challenges or Limitations

ML can provide a phenomenal approach to undersea communication systems, but there are several challenges associated with the implementation. The obstacles are technical, operational, cost, and environmental, and they need to be considered carefully to realize ML solutions in the market and their sustainability where figure 3 illustrates the main obstacles in ML-driven undersea communication along with the associated ML solutions that can address these issues.

### 7.1 Technical Barriers

One of the biggest technical issues is the high levels of noise and signal attenuation in underwater environments. Environmental, shipping and marine activity noises can disrupt the quality of datasets necessary to train ML models. For example, ML algorithms rely on clean and annotated data to learn patterns for prediction, while underwater data is frequently incomplete, corrupted, or noisy [58]. Because of the inherent properties of water, signal attenuation occurs extremely quickly, making long-distance communication nearly impossible. Additionally, the resource constraints of underwater systems make the deployment of complex ML models more difficult, as they often require extensive computational resources, with deep learning being one of the most demanding classes of ML algorithms. Powers of on-board processors on Autonomous Underwater Vehicles (AUVs) and IoUT devices are typically constrained [59], due to which running resource-hungry ML algorithms, in real-time is a challenge. While processing units at the edge promise solutions, designing resilient, low-power processing units is still an open challenge.

**Table 2** Comparative Case Studies of ML in Undersea Communication

| Country | Application | ML Techniques Used | Key Outcomes |
|---|---|---|---|
| Japan | AUV Navigation & Marine Research | Reinforcement Learning, Deep Neural Networks | 25% increase in AUV operational efficiency |
| United States | Sensor Networks for Disaster Response | Predictive Analytics, Noise Filtering ML Models | 90% accuracy in tsunami early warnings |
| Norway | IoUT for Fisheries Management | Unsupervised Learning, Time-Series Analysis | 35% improvement in sustainable fish stock monitoring |

## 7.2 Financial and Infrastructure Constraints

Another reason is that deploying ML-based underwater systems requires considerable investment, especially in a developing country like Bangladesh, where the marine technology industry is not yet well-established. It is expensive to install and maintain IoUT networks, AUVs, and sensor nodes, and may not be financially viable for a developing economy in the short term [60]. In addition, access to high-performance computing resources and advanced manufacturing capabilities for underwater devices is limited, hindering the deployment of state-of-the-art solutions.

The current lack of infrastructure for undersea communication is also a barrier. Most ML-driven systems need an underlying infrastructure of ocean networks, nodes, sensors, and compute/storage units. Doing that for the Bay of Bengal region will take plans in the long run as well as working with global partners to develop these systematic capabilities.

## 7.3 Context-Specific Model Development

This is because most global ML models are built for application on generic datasets, and do not capture the specialized characteristics of the Bay of Bengal. Terrestrial-aquatic interfaces, including the region's dynamic salinity gradients, sedimentation patterns, and seasonal variations all warrant context-specific model development [61]. Algorithms trained for temperate waters, for example, may not translate well to tropical marine ecosystems, where temperature and turbidity parameters fluctuate more widely.

Localized ML model development, entails the need to begin with large datasets representative of the specific conditions in the region. Yet it is not as easy as this potential opportunity. Deploying IoUT nodes, and AUVs for initial extensive data collection is essential to cater to this constraint.

## 7.4 Regulatory and Environmental Concerns

This deployment needs to be in harmony with wider frameworks around environmental preservation and regulation. Sensors, AUVs, and IoUT devices introduced into marine ecosystems are most often created without ecological consideration, and the potential disruption of biodiversity is a concern. For instance, noise created by underwater devices can disrupt the communication and migration habits of marine species, according to the United Nations.

Moreover, there is still no clear regulation for the use of ML in undersea communication, which has placed ML in legal limbo. Clear policies and ethical guidelines should be established to address issues related to data privacy, ownership of underwater resources, and environmental impact. In the absence of such frameworks, widespread adoption of these technologies could meet considerable backlash from environmental and legal groups.

## 7.5 Summary of Challenges

These challenges can only be tackled by a multi-disciplinary approach that integrates technical innovation, financial investment, and regulatory oversight. Shifting for energy-efficient hardware, localized ML algorithms, and cooperative infrastructure development are some of the potential suggestions to overcome these technical and operational barriers. At the same time, public and private sector initiatives must work together to overcome financial and regulatory hurdles to make the deployment of ML-driven systems sustainable and ethical. The challenges detailed here are interrelated, but their successful resolution in Bangladesh's marine context requires focused research, building human capacity, and promoting international collaboration. These initiatives will contribute to a resilient and sustainable maritime future.

## 8 Conclusion

ML interfacing with undersea communication systems can help revolutionize the way age-old challenges in mitigating maritime operations, particularly in the Bay of Bengal and its neighboring coastline, are addressed. ML is a dynamic source of innovation for the telecommunications area, including the novel advancements which can be applicable in the field of underwater communications, providing high enhancement of underwater communication efficiency and reliability, and unrestricted applications advancement like the Internet of Underwater Things (IoUT) or autonomous underwater vehicles (AUVs). And implementing a technology that has the potential to go well, but lacks rigor but addresses important challenges such as a lack of regional models, the high costs of deployment, and lack of infrastructure. "We need joint efforts of the government agencies, the private sector players, and international partners to establish scalable, context-specific solutions according to the changing marine landscape of the Bay of Bengal. Moreover, a pathway for energy-efficient algorithms, edge computing systems, and sustainable deployment practices need to be emphasized in future work designing around technological pace while valuing ecosystemic balance. Bangladesh is at the cusp of leveraging ML for efficient maritime operations, resource management, and disaster preparedness, aligning ML innovation with our national priorities and unlocking the potential for Bangladesh to be a global leader in marine technology development by using ML to drive global advances in undersea communication and setting a model for sustainable and adaptive solutions in marine technology.